\documentclass[letterpaper, 10 pt]{ieeeconf}\IEEEoverridecommandlockouts
\usepackage{cite}
\usepackage{amsmath,amssymb,amsfonts}
\usepackage{algorithmic}
\usepackage{graphicx}
\usepackage{textcomp}
\usepackage[dvipsnames]{xcolor}

\usepackage{customCommands}
\usepackage{comment} 
\usepackage{url}

\usepackage[bookmarks,%
			colorlinks = true,%
			linkcolor = blue,%
			citecolor = red,%
			allcolors = black,%
			pdfauthor = {Joan\ Sol\`a},%
			pdftitle = {Wolf:\ a\ modular\ estimation\ framework\ for\ robotics\ based\ on\ factor\ graphs},%
			pdftex
			]{hyperref}

\newcommand{\wolf}{{\sc Wolf}}
\newcommand{\ros}{{\sc ROS}}

\newcommand{\node}[1]{\texttt{#1}}
\newcommand{\problem}{\node{Problem}}
\newcommand{\hardware}{\node{Hardware}}
\newcommand{\sensor}{\node{Sensor}}
\newcommand{\sensors}{\node{Sensors}}
\newcommand{\processor}{\node{Processor}}
\newcommand{\processors}{\node{Processors}}
\newcommand{\trajectory}{\node{Trajectory}}
\renewcommand{\frame}{\node{Frame}}
\newcommand{\frames}{\node{Frames}}
\newcommand{\capture}{\node{Capture}}
\newcommand{\captures}{\node{Captures}}
\newcommand{\feature}{\node{Feature}}
\newcommand{\features}{\node{Features}}
\newcommand{\factor}{\node{Factor}}
\newcommand{\factors}{\node{Factors}}
\newcommand{\map}{\node{Map}}
\newcommand{\landmark}{\node{Landmark}}
\newcommand{\landmarks}{\node{Landmarks}}

\newcommand{\etc}{\emph{etc}}

\renewcommand{\secRef}[1]{Sec.\,\ref{#1}}

\newcommand{\core}{\node{core}}

\newcommand{\rev}{\color{blue}}
\renewcommand{\rev}{}

\begin{document}


\title{\centering \wolf: A modular estimation framework\\ for robotics based on factor graphs}
\author{%
Joan Sol\`a$^{*}$, Joan Vallv\'e$^{*}$, Joaquim Casals, J\'er\'emie Deray,\\ M\'ed\'eric Fourmy, Dinesh Atchuthan, Andreu Corominas-Murtra and Juan Andrade-Cetto
\thanks{* These authors contributed equally.}
\thanks{JS, JVN, JC, JD, ACM and JAC are/were with the Institut de Rob\`otica i Inform\`atica Industrial (IRI), CSIC-UPC, Llorens Artigas 4-6, Barcelona (Corresponding author e-mail: jsola@iri.upc.edu).}
\thanks{JD was also with PAL Robotics, Pujades 77, Barcelona.}
\thanks{MF and DA are with LAAS-CNRS, 7 Av. Colonel Roche, Toulouse.}
\thanks{This work was partially supported by the EU H2020 projects LOGIMATIC (Galileo-2015-1-687534), GAUSS (Galileo-2017-1-776293), TERRINET (INFRAIA-2017-1-730994), and MEMMO (ICT-25-2017-1-780684); and by the Spanish State Research Agency through projects EB-SLAM (DPI2017-89564-P), EBCON (PID2020-119244GB-I00), and the Mar\'ia de Maeztu Seal of Excellence to IRI (MDM-2016-0656).}
}

\maketitle



\begin{abstract}

This paper introduces \wolf, a C++ estimation framework based on factor graphs and targeted at mobile robotics. 
\wolf\ can be used {\rev beyond SLAM} to handle self-calibration, model identification, or the observation of dynamic quantities other than localization. 
{\rev The architecture of \wolf\ allows for a modular yet tightly-coupled estimator.}
Modularity is {\rev enhanced via} reusable plugins that are loaded at runtime {\rev depending on application setup.
This setup} is achieved conveniently through YAML files, allowing users to configure a wide range of applications without the need of writing or compiling code.
Most {\rev procedures} are coded as abstract algorithms in base classes with varying levels of specialization. 
Overall, {\rev all} these assets allow for coherent processing and favor code re-usability and scalability.
{\rev \wolf\ can be used with \ros, and is made publicly available and open to collaboration.}
    
\end{abstract}

\section{Introduction}
\label{sec:introduction}

State estimation is key in many complex dynamic systems when it comes to controlling them. 
In robotics, and especially for robots such as humanoids, legged robots or aerial manipulators, this is of special importance because of their high dynamics, inherent instability, and the necessity of entering in contact with the environment. 
This estimation includes the robot states at high rate and low lag, important for control, but also and crucially the environment, necessary for motion planning and interaction. 
Since the formalism of simultaneous localization and mapping (SLAM), which tackles both sides in a unified way, robot state estimation has evolved enormously and now reaches a level of complexity that goes well beyond the SLAM paradigm,  tackling aspects such as sensor self-calibration, geo-localization, system identification, or the estimation of dynamic quantities other than localization.

With the increase of real and complex robotic applications of very diverse nature (2D, 3D, wheeled, legged, flying, submarine), the problems of integration, modularity, code re-usability, scalability and the likes are becoming real challenges that demand solutions. 
The issue then is in finding appropriate software architectures able to handle such variety in a practical way.
Building such an estimation system requires the adoption of a number of design decisions affecting both the way the estimator works internally and the way this is used by developers and practitioners.

One of such decisions involves the nature of the estimation engine. 
Historically, the Bayesian filtering paradigm has been used for sensor fusion and SLAM, with the Extended Kalman Filter or more modern variants.
In RT-SLAM \cite{ROUSSILLON-11a} we fused vision, IMU and GNSS using EKF. 
Recent examples are MaRS\cite{brommer-21-mars} and  PRONTO \cite{camurri-20-pronto}, this second one used for legged robots. 
With the advent of powerful computers and efficient nonlinear solvers, it is becoming standard to switch to nonlinear iterative maximum a-posteriori (MAP) optimization, particularly in the form of factor graph optimizers solved by least-squares \cite{dellaert-17-gtsam,GRISETTI-10-GRAPHSLAM,ILA-17_SLAM++}, which achieve better performances in most fronts \cite{STRASDAT-10} including speed and accuracy.

\begin{figure}[t]
  \centering
  \includegraphics
  [width=0.9\linewidth]
  {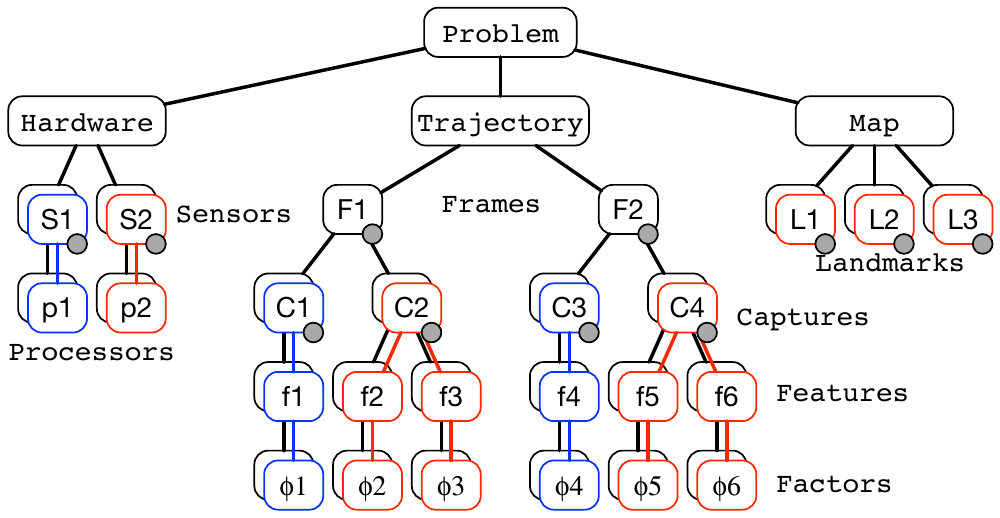}
  \caption{The \wolf\ tree, showing the top node \problem\ with branches \hardware, \trajectory\ and \map. Children nodes shown are: \sensors~(S) and their associated \processors~(p), trajectory \frames~(F), \captures~(C) of raw data, detected \features~(f), their corresponding \factors~($\phi$), and the \landmarks~(L) added to the \map.  Base classes, in black, are contained in \wolf\ \core. Classes derived from them, in color, are packaged in separate libraries called \wolf\ plugins. The displayed tree corresponds to a visual-inertial setup, comprising an IMU plugin (blue) and a vision plugin (red).   Each plugin comes with proper sensor, processor, capture, feature, factor, and landmark types. Classes with state-blocks are marked with a gray dot. See the accompanying video.}
  \label{fig:wolf-tree}
\end{figure}

A second aspect concerns loosely versus tightly coupled estimators.
In loose coupling, a set of independent modules, each accepting the data from one (or a small subset of) sensor, produce state estimates of the system. 
These are then fused in a second stage to find a unique weighted estimate. 
Loose coupling as in \eg~\cite{scona-17-semidense-humanoid} is practical because it is easy to develop and easily scales up, for example taking advantage of the \ros\ architecture.
However, it does not offer great performances, due to important cross-correlations being ignored, preventing \eg\ sensor self-calibration.
In contrast, tightly coupled systems can enable the observation of otherwise hidden states such as intrinsic and extrinsic sensor parameters, thus improving the overall accuracy. 
Because of the need for maintaining cross-correlations, sometimes these systems exhibit no modularity, although this can be circumvented as explained below.

Modularity is precisely a third major design criterion. 
Non-modular estimators such as ORB-SLAM \cite{campos-21-orbslam3} or Cartographer \cite{hess-16-cartographer} can achieve very high performance, but tend to be specialized to a particular sensor setup. Modifying or scaling them up is difficult, since one needs to be aware of all the intricate inter-relations.
Modular systems improve versatility, scalability and re-usability, and can more easily be made to withstand sensor failure, greatly improving robustness. 
However, modularity can come at the price of compromising  performance, since communication between modules is often nonexistent, preventing potential synergies between them.

A very interesting compromise amongst all these dichotomies can be achieved with modern architectures, pioneered by Klein's PTAM \cite{KLEIN-08}, that divide the  estimator in (a set of) front-ends and one back-end. This philosophy quickly evolved to the graph-SLAM paradigm \cite{fuse-stack,blanco-19-MOLA,labbe-19-rtab-map,colosi-20-plugandplay}, where the element of union between front-ends and back-end is a factor graph, which is a probabilistic representation of the full problem that is solved iteratively using off-the-shelf nonlinear solvers.
This offers some key advantages.
The full representativeness of the graph enables tightly coupled estimation. 
Modularity is achieved by assigning one front-end to every sensor, each contributing its states and/or factors to the graph.
Establishing the right frontiers between front-ends, and the interface with the graph and back-end, allows us to set up many different systems by combining re-usable parts that have been developed separately.
Packaging these parts in separate libraries or \emph{plugins} allows independent and decentralized development, thus enabling scalable open-source projects to emerge.
This is necessary to ultimately standardize fusion for robotics in practice so that it can be used by non-experts.

\paragraph{Closely related works}
This idea above is the main motivation behind \wolf\ and is shared by a few other recent projects.
They differ in the way the components and algorithms are organized, something that, beyond performance, impacts the way developers and users interact with each system.
Perhaps the simplest example is the Fuse stack \cite{fuse-stack}, with sensor and motion front-ends organized in plugins and devoted to contribute factors to the graph directly, and a number of \ros\ publishers to output the results of the optimization.
Unfortunately, Fuse has not been published as a scientific communication and has little documentation, making it difficult to evaluate.
Similar proposals are MOLA~\cite{blanco-19-MOLA}, which has been reportedly tested only with LIDAR and wheel odometry sensors, and RTAB-Map~\cite{labbe-19-rtab-map} which bases loop closure on processing RGBD images.
A slightly better approach in our opinion is {Plug-And-Play SLAM}~\cite{colosi-20-plugandplay}, a SLAM architecture in C++ that realistically aims at standardizing multi-modal SLAM. 
The architecture is organized around some SLAM-driven abstract design patterns, which form the core module.
New sensor modalities can be added from other modules by editing a configuration file.
P\&P-SLAM has been demonstrated with wheel odometry, IMU, LIDAR and RGBD vision.

\paragraph{Contributions} 
Our proposal \wolf\ offers some key differences over the aforementioned systems. 
First, we use an intermediate layer between the front-ends and the graph, called the \wolf\ tree.
This stores all the information in a comprehensive way, mimicking the elements encountered in the real problem. 
That is, our architecture is mainly constituted by objects and not by procedures like P\&P-SLAM.
The \wolf\ tree is completely accessible as all links can be traversed bidirectionally. 
This allows the front-ends to make use of precious processing information generated by other front-ends, giving them the possibility to benefit each other while keeping the architecture completely modular. 
The \wolf\ tree can be printed at any time and becomes a powerful book-keeping tool for tuning and debugging.
Second, and in a similar spirit as P\&P-SLAM, we provide the most common algorithms in abstract form (motion pre-integration, feature or landmark tracking, and loop closing), making the generation of specific algorithms easier. 
Third, this abstraction allows us to include sensor self-calibration natively, both for extrinsic and intrinsic parameters. 
Crucially, we offer generic calibration for sensors requiring pre-integration (odometers, IMU, force/torque).
\wolf\ also handles time-varying sensor parameters, such as IMU biases, natively and generically.
Finally, we showcase real applications in both 2D and 3D setups: IMU, wheel odometry, LIDAR, vision (in several ways), force and torque, limb encoders, and GNSS (in several ways), overall comprising a larger variety of sensors than any of the previous proposals.

\wolf\ is made available freely for research purposes and comes with extensive documentation.\footnote{Documentation, access to all repositories and installation instructions can all be found at \url{http://www.iri.upc.edu/wolf}.}
We encourage the reader to also watch the accompanying video to complement or clarify the information provided in this paper.


\section{Architecture}
\label{sec:architecture}

\subsection{Overview}

{\rev 
One of the key originalities of \wolf\  is its central data structure, called the \wolf\ tree (\figRef{fig:wolf-tree}, \secRef{sec:wolf-tree}). 
The \wolf\ tree is a tree of abstract base classes reproducing the elements of the robotic problem (\sensors, \processors, \frames, \features, \factors, \etc.). 
To account for different sensing or processing modalities,
these classes are derived to implement specific functionalities. 
These elements can be added to the system in a modular way, allowing great flexibility to setup different applications.
On top of this modularity, scalability is achieved by splitting the project into independent libraries called \emph{plugins}, each containing a set of derived classes corresponding to one of such modalities (\eg\ laser, vision, imu, and others). 

\wolf\ \processors, which are part of the \wolf\ tree, constitute the estimator's front-ends (\secRef{sec:processors}).
As the back-end, \wolf\ interfaces with one graph-based solver (\secRef{sec:solver}).
In brief, incoming raw data is processed by the processors to populate the \wolf\ tree, which is translated to a factor graph that is repeatedly solved. 

To configure a particular application, we designed a fully automated setup mechanism based exclusively on YAML files describing the full robotic setup (\secRef{sec:autoconf}). 
For convenience, \wolf\ can be interfaced with \ros\ (\secRef{sec:ros-integration}). 

}


\subsection{The \wolf\ tree}
\label{sec:wolf-tree}

\wolf\ organizes the typical entities appearing in robot state estimation in a tree structure (\figRef{fig:wolf-tree}), with nodes for:

\begin{itemize}
  \item 
the robot's \hardware\ consisting of all of its \sensors, and their \processors\ of raw data;

  \item 
the robot's \trajectory\ over time: \frames, \captures\ of raw data, \features\ or metric measurements extracted from that data, and \factors\ relating these measurements to the state blocks of the system; 
  \item 
the \map\ of \landmarks\ in the environment.

\end{itemize}
Most of these nodes are basically data holders: \sensor\ (extrinsic and intrinsic parameters), \frame\ (robot state), \capture\ (raw sensory data), \feature\ (metric measurement) and \landmark\ (state, appearance descriptor). 
Nearly all the processing work resides in three nodes: \processors\ (at the front end), \factors\ (at the back end), and \problem\ (tree management).

\begin{figure}[t]
\centering
\includegraphics[width=0.9\linewidth]{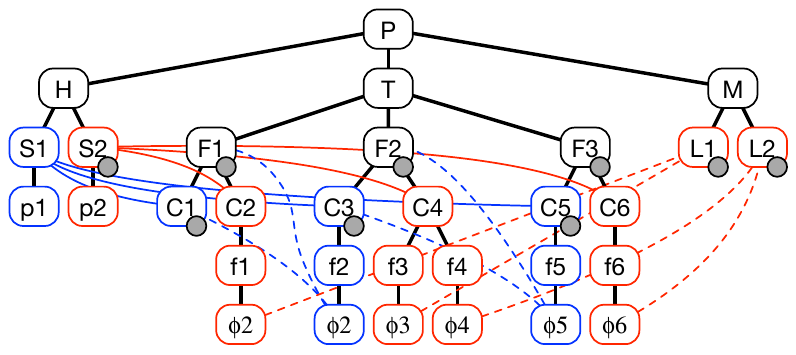}
\caption{Pointers across branches allow trees in the likes of \figRef{fig:wolf-tree} to be converted to a factor graph. 
\captures\ have always a pointer (solid) to the \sensor\ that created them. 
\factors\ have pointers (dashed) to nodes of the tree that appear in their measurement model. 
The tree shows nodes and measurements related to vision (red) and IMU (blue). 
This setup allows for self-calibration of the camera's extrinsic parameters (state blocks in S2) and the tracking of the IMU biases (in IMU captures C1, C3, C5). 
See \figRef{fig:wolf-graph} for the resulting factor graph.}
\label{fig:wolf-tree-graph}
\end{figure}

In each branch of the tree, \wolf\ nodes may have any number of children. 
{\rev
All links are bidirectional making the tree fully accessible from any point. 
}
Crucially, some key connections break the tree structure and provide the basis for the creation of a meaningful network of relations describing the full robotic problem (\figRef{fig:wolf-tree-graph}). 
These connections are: 
(a) \captures\ have access to the \sensor\ that created them; 
and (b) \factors\ have access to each node containing state blocks necessary to compute the factor residual. 
This richer network is easily interpreted as a factor graph (\figRef{fig:wolf-graph}), which is solved by the solver or back-end.

At startup the \wolf\ tree is initialized with the full definition of the \hardware\ branch and, optionally, a first \frame\ with the robot's initial condition and/or a pre-defined \map\ of the environment.
This information is extracted at run-time from YAML configuration files, see \secRef{sec:autoconf}.

\subsection{Front-ends: Processors}
\label{sec:processors}

The rest of the \wolf\ tree is incrementally built by the \processors\ as data is being gathered. 
\processors\ are responsible for many things:
extracting \features\ from raw data in \captures;
   associating \features\ with other nodes in the tree;
   creating \frames\ (and conditionally to this, creating \factors\ and new \landmarks);
  joining \frames\ created by other \processors; 
  integrating motion; and
  closing loops.
%
See \secRef{sec:processor_algorithms} for further details.

\subsection{\wolf\ tree manager}
\label{sec:tree-manager}

Since \processors\ only populate the tree with new nodes, a manager that is able to remove nodes is necessary to keep the problem at a controlled size. 
The tree manager is a member of the \problem\ node. 
We provide with \core\ two managers implementing sliding windows of \frames. 
Other managers performing \eg\ graph sparsification \cite{vallve-18-sparsification-RAL} should be implemented by deriving from a base class.

\subsection{Back-end: graph and solver}
\label{sec:solver}

The back-end can be explained by means of a factor graph, with its state blocks and factors, and a graph solver. 

\begin{figure}[t]
\centering
\hfill
\includegraphics[height=0.32\linewidth]{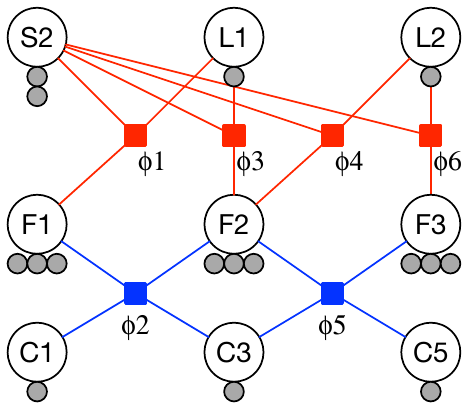}
\hfill
\includegraphics[height=0.32\linewidth]{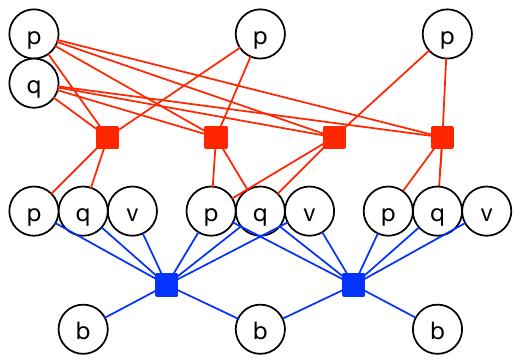}
\hfill
\caption{Two flavors for the factor graph derived from the tree in \figRef{fig:wolf-tree-graph}. \emph{Left}: human-readable graph: nodes are \wolf\ nodes. State blocks in gray, and factors in color (red: vision, blue: IMU). \emph{Right}:  graph as seen by the solver: nodes are state blocks ($p$:~position, $q$:~orientation, $v$:~velocity, $b$:~bias).}
\label{fig:wolf-graph}
\end{figure}

\subsubsection{State blocks}
\label{sec:state-blocks}

\begin{figure}[t]
\centering
\includegraphics[height=0.22\linewidth]{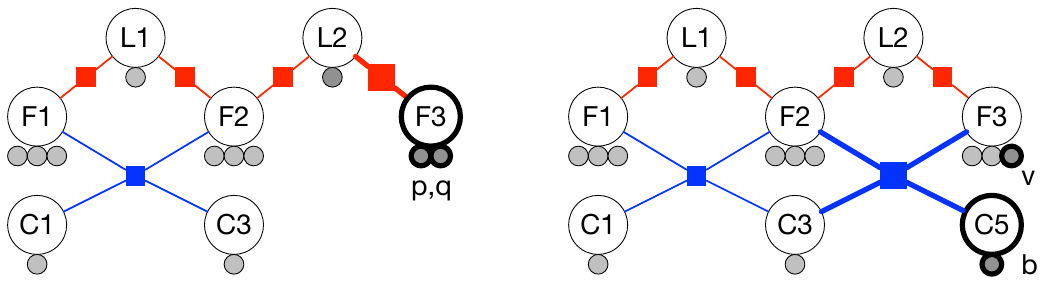}
\caption{{\rev Dynamic evolution of the \frame's structure of state blocks in a visual-inertial setup.} \emph{Left}: The visual processor adds a new \frame\ F3 with state blocks $p,q$ necessary for the new landmark observation factor (thick red). \emph{Right}: Then, the IMU \processor\ joins F3 (see Sec.~\ref{sec:kf-join}) and adds to it a new state block $v$ necessary for the pre-integrated IMU factor (thick blue). 
It also appends its IMU \capture\ C5 with the bias block $b$.}
\label{fig:new-kf}
\end{figure}

Some nodes in the tree (\sensor, \frame, \capture\ and \landmark) contain instances of the state block class. 
A state block is the minimum partition of the full problem state that carries a particular meaning (position, orientation, sensor parameters, etc.). 
Crucially, state blocks in \frames\ can be added dynamically (\figRef{fig:new-kf}). 
This allows for \processors\ and \factors\ requiring a different set of state blocks for a given \frame\ to work together seamlessly.

A state block $i$ contains a state vector $\bfx_i$ and, optionally, a local (or tangent) parametrization $\Delta\bfx_i$ used to perform estimation on the manifold of $\bfx_i$. 
%
A state block can be fixed by the user, meaning that its values are treated as fixed parameters, and hence not optimized by the solver. 

\subsubsection{Factors}
\label{sec:factors}

\factors\ are responsible for computing a residual $\bfr(\bfx_1,\dots,\bfx_N)$ each time the solver requires it. 
%
Each \factor\ contains pointers to all the state blocks $\bfx_1,\dots,\bfx_N$ appearing in the observation model.
\factors\ can optionally include a loss function allowing for robust estimation in front of outliers. 
Since the solver requires the Jacobians of $\bfr$ to perform the optimization, we provide base \factors\ prepared to produce analytical or automatic Jacobians. 

\subsubsection{Solver}


We designed \wolf\ to be used with nonlinear graph solvers. 
These are available as independent libraries such as Google Ceres \cite{ceres-solver}, GTSAM \cite{dellaert-17-gtsam}, g2o \cite{kummerle-11-g2o}, SLAM++ \cite{ILA-17_SLAM++}, and others.
%
Interfacing \wolf\ with the solver is
relegated to solver wrappers, which are encapsulated in classes deriving from a base class in \wolf\ \core. 
{\rev
The wrapper essentially makes the \wolf's state-blocks and \factors\ accessible to the solver by bridging the API on both sides.
}
We provide in \core\ a wrapper to Google Ceres.

\subsection{Modular development through independent plugins}

{\rev
As explained already, \wolf\ consists of both abstract classes and specialized classes.
To promote modularity and scalability, \wolf\ encapsulates all abstract material in  \wolf\ \core\ while specializations are packaged in other libraries called \wolf\ plugins.
Each plugin corresponds typically to a certain sensor modality, or to a certain way to process a sensor modality's data. 
Plugins are meant to be reused across different applications.
New plugins can be added in case a certain sensing modality or processing method does not exist amongst the available plugins.
These new plugins can be incorporated to the \wolf\ general repository, or kept elsewhere, preferably available for the community.
}
Currently \wolf\ includes plugins for \texttt{vision}, \texttt{apriltag}, \texttt{gnss}, \texttt{imu}, \texttt{laser} and \texttt{bodydynamics}, this last one devoted to the estimation of the whole-body dynamics of articulated robots such as humanoids, quadrupeds or aerial manipulators.
{\rev
Plugins can depend on other plugins.
For example, the \texttt{apriltag} plugin depends on \texttt{vision} and contributes a new way of processing visual information.
}

\subsection{Multi-threading}

There is one dedicated thread for the solver and one for each \wolf\ \ros\ publisher (see \secRef{sec:ros-integration}). 
At the moment, all processors are executed in one single other thread ---see \secRef{sec:conclusions} for planned future improvements.


\renewcommand{\node}[1]{\lowercase{#1}}

\begin{figure}[t]
\centering
\includegraphics[width=0.9\linewidth]{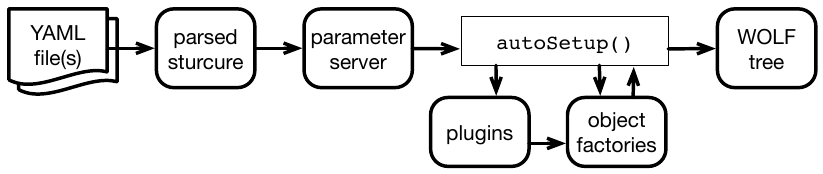}
\caption{\wolf\ auto-configuration pipeline. 
YAML files are parsed and converted to a parameter server. 
The method \texttt{Problem::autoSetup()} reads from this server and loads the plugins, whose object creators are all registered in the factories in \texttt{core}. 
These are then used in \texttt{autoSetup()} to create all objects to initialize the \wolf\ tree.
}
\label{fig:autoconf}
  \vspace{-8pt}
\end{figure}

\section{Features}
\label{sec:features}

\subsection{Full auto-configuration from YAML files}
\label{sec:autoconf}

\wolf\ includes a mechanism for providing all configuration parameters to the \wolf\ problem through YAML files.\footnote{See \url{https://gitlab.iri.upc.edu/mobile_robotics/wolf_projects/wolf_ros/demos/wolf_demo_laser2d/-/blob/devel/yaml/demo_laser2d.yaml} for an example of YAML file describing a real robotics problem.}
These files contain: (a) the \sensors\ and their parameters; (b) the \processors\ and their parameters; (c) the solver and its parameters; (d) the tree manager and its parameters; (e) the initial robot state; and optionally (f) the initial \map.

\figRef{fig:autoconf} shows the auto-configuration process: starting from a set of YAML files, a parser and a parameter server are used in conjunction with object factories to create the \wolf\ tree and leave the robot ready to start its mission.
%
\wolf\ automatically loads the plugins at run-time and registers all derived object creators in dedicated factories in \texttt{core}. 
This means that{, \rev having the necessary plugins available,} different estimation problems can be set up without the need for writing and compiling code, but just specifying the problem setup at YAML level. 
See the accompanying video for a demonstration.

\subsection{Sensor parameters management}

Sensor parameters (intrinsic and extrinsic) 
can be \emph{static} (\ie, constant ---not to be confused with \emph{fixed}, see \ref{sec:state-blocks}) or \emph{dynamic} (\ie, time-varying).
Sensor parameters declared static are stored in state blocks in the \sensor\ object itself.
Dynamic sensor parameters are stored in state blocks in the \captures\ created by the \sensor\ so that they can be associated with the particular timestamp of the \capture\ (and therefore also of its parent \frame).

Sensor parameters that are to be estimated require the state blocks to be unfixed (see \ref{sec:state-blocks}). 
If these are static, we speak of \emph{parameter calibration}. If they are dynamic, we speak of \emph{parameter tracking}.  
For example, see Figs.~\ref{fig:wolf-tree-graph} and \ref{fig:wolf-graph}, IMU biases are varying with time and need to be continuously tracked. Instead, camera extrinsic parameters are constant and may just require calibration.

\subsection{Abstract (or generic) processor algorithms}
\label{sec:processor_algorithms}

The variability of the processing algorithms may be very large. 
However, there are three well-identified groups of algorithms in SLAM: motion integration, feature tracking, and loop closing. 
Their basic functionalities, coded in respective base classes, are explained below.

\subsubsection{Motion pre-integrators}
\label{sec:proc_motion}

This processor performs generic motion pre-integration with automatic sensor calibration or tracking. 
It is based on {\rev our} generalized pre-integration theory {\rev \cite[Section 4.3]{atchuthan-18-thesis}} implementing the following pipeline of operations (notation: 
$\bfu$: raw motion data; 
$\bfv$ calibrated motion data; 
$\ol\bfc$: calibration parameters' initial guess; 
$\delta$: current motion delta; 
$\ol\Delta$: pre-integrated motion delta; 
$\bfJ^y_x\triangleq\partial y/\partial x$: Jacobians computed according to Lie theory \cite{SOLA-18-Lie}; 
$\bfQ_x$: covariance of $x$):
\begin{itemize}
\item
Pre-calibrate motion data ~~~$\bfv = f(\bfu,\ol\bfc),~\bfJ^\bfv_\bfu,~\bfJ^\bfv_\bfc$.
\item
Compute current delta ~~~~~~\,$\delta = g(\bfv),~\bfJ^\delta_\bfv$.
\item
Pre-integrate delta ~~~~~~~~~~$\ol\Delta\gets\ol\Delta\circ\delta,~\bfJ^\Delta_\Delta,~\bfJ^\Delta_\delta$.
\item
Integrate covariance \\$\bfQ_\Delta \gets \bfJ^\Delta_\Delta\,\bfQ_\Delta\,{\bfJ^\Delta_\Delta}\tr+\bfJ^\Delta_\delta\bfJ^\delta_\bfv\bfJ^\bfv_\bfu\,\bfQ_\bfu\,(\bfJ^\Delta_\delta\bfJ^\delta_\bfv\bfJ^\bfv_\bfu)\tr$.
\item
Integrate Jacobian \wrt calibration parameters \\$\bfJ^\Delta_\bfc \gets \bfJ^\Delta_\Delta\bfJ^\Delta_\bfc +\bfJ^\Delta_\delta\bfJ^\delta_\bfv\bfJ^\bfv_\bfc$.
\end{itemize}

\wolf\ can produce high throughput estimates at sensor rates up to the kHz range, which can be used {\rev by other processors} and for feedback control of highly dynamic robots such as humanoids, quadrupeds or aerial manipulators. 
These states are computed with $\bfx_t = \bfx_i\boxplus\ol\Delta_{it}$, where $\bfx_i$ is the last frame at time $i\!<\!t$ and $\ol\Delta_{it}$ is the delta pre-integrated from times $i$ to $t$.
Classes deriving from \texttt{ProcessorMotion} only have to implement the function $g(f(\bfu,\bfc))$, the delta composition $\circ$ and the operator $\boxplus$.


\subsubsection{Trackers}

They extract features from raw data such as images or laser scans and track them over time. We offer two abstract variants. One associates \features\ in the current \capture\ with \features\ in the \capture\ at the last \frame. The second one associates \features\ with \landmarks\ in the \map\ and is able to create new \landmarks.

\subsubsection{Loop closers}

They search for \frames\ in the past having \captures\ that are, in some robust sense, similar to the present one. Then, based on some geometrical analysis of the data, they establish the appropriate \factors\ between the two sensor \frames.

\subsection{Sensor self-calibration}

{\rev
We distinguish three self-calibration strategies, all logically subject to the observability conditions given by each particular sensor setup. 
First, the calibration of sensor extrinsic parameters just requires the factors to account for the extra robot-to-sensor transformations, which we provide. 
Second, intrinsic self-calibration requires the factors' observation model to account for such parameters: they must be specifically coded in each case and cannot be generalized. 
Finally, the most difficult intrinsic self-calibration for sensors requiring pre-integration is also part of our generalized pre-integration theory {\rev \cite[Sec.\,4.3]{atchuthan-18-thesis}}, see \secRef{sec:proc_motion}.
At the \factor\ side, the pre-integrated delta $\ol\Delta$ is corrected for values of the calibration parameters $\bfc$ different from the initial guess $\ol\bfc$ following the linearized formula $\Delta(\bfc) = \ol\Delta\op\bfJ^\Delta_\bfc(\bfc-\ol\bfc)$. 
The resulting residual $\bfr(\bfx_i,\bfx_j,\bfc)=\bfQ_\Delta^{-\top/2}(\Delta(\bfc)\om(\bfx_j\boxminus\bfx_i))$ between consecutive frames $i,j$ enables the observation of the sensor parameters $\bfc$. 
See \cite{LUPTON-09, forster2017-TRO} for seminal works on motion pre-integration for the IMU ---notice that $\bfc$ are biases in the IMU case but can be any sensor parameter, static or dynamic, in the general case. 
Examples of application of this method are \cite[Chap.~3]{atchuthan-18-thesis}, \cite{DERAY-19-SELFCALIB,fourmy-19-april}, and also in \secRef{sec:experiments}.
}

\subsection{Automatic frame synchronization}
\label{sec:frame-sync}

In multi-sensor setups, efficient frame creation policies can be tricky. We designed a decentralized strategy where each processor may independently decide to create a frame. Once created, the frame is broadcasted for other processors to join (see \figRef{fig:new-kf}). This technique requires all sensory data to be timestamped by a common clock, or at least by clocks that are properly synchronized.

\subsubsection{Frame creation and broadcast}

Each processor defines its own frame creation policy. 
Besides trivial policies like time elapsed or distance traveled, frames can be more smartly distributed if the decision is made based on the processed data. 
For example, a \feature\ tracker may create a \frame\ when the number of feature tracks since the last frame has dropped below a threshold. 
Frame creation is followed by appending the \capture\ with its \features, and creating the appropriate \factors\ (and \landmarks, if any).
Once created, \frames\ are broadcasted for other processors to join.

\subsubsection{Frame join}
\label{sec:kf-join}

Broadcasted \frames\ are received by all other \processors. On reception, processors will compare their \capture's timestamps with the timestamp of the received \frame. A check of the time difference against the processor time tolerances is performed. 
If passed, the processor adds, if necessary, state blocks to the \frame\ (see \ref{sec:state-blocks} and \figRef{fig:new-kf}), appends the selected \capture\ as a new child of the received \frame, and creates the appropriate \features\ and \factors\ (and \landmarks, if any).


\section{\ros\ integration}
\label{sec:ros-integration}

\begin{figure}
  \centering
  \includegraphics[width=0.9\linewidth]{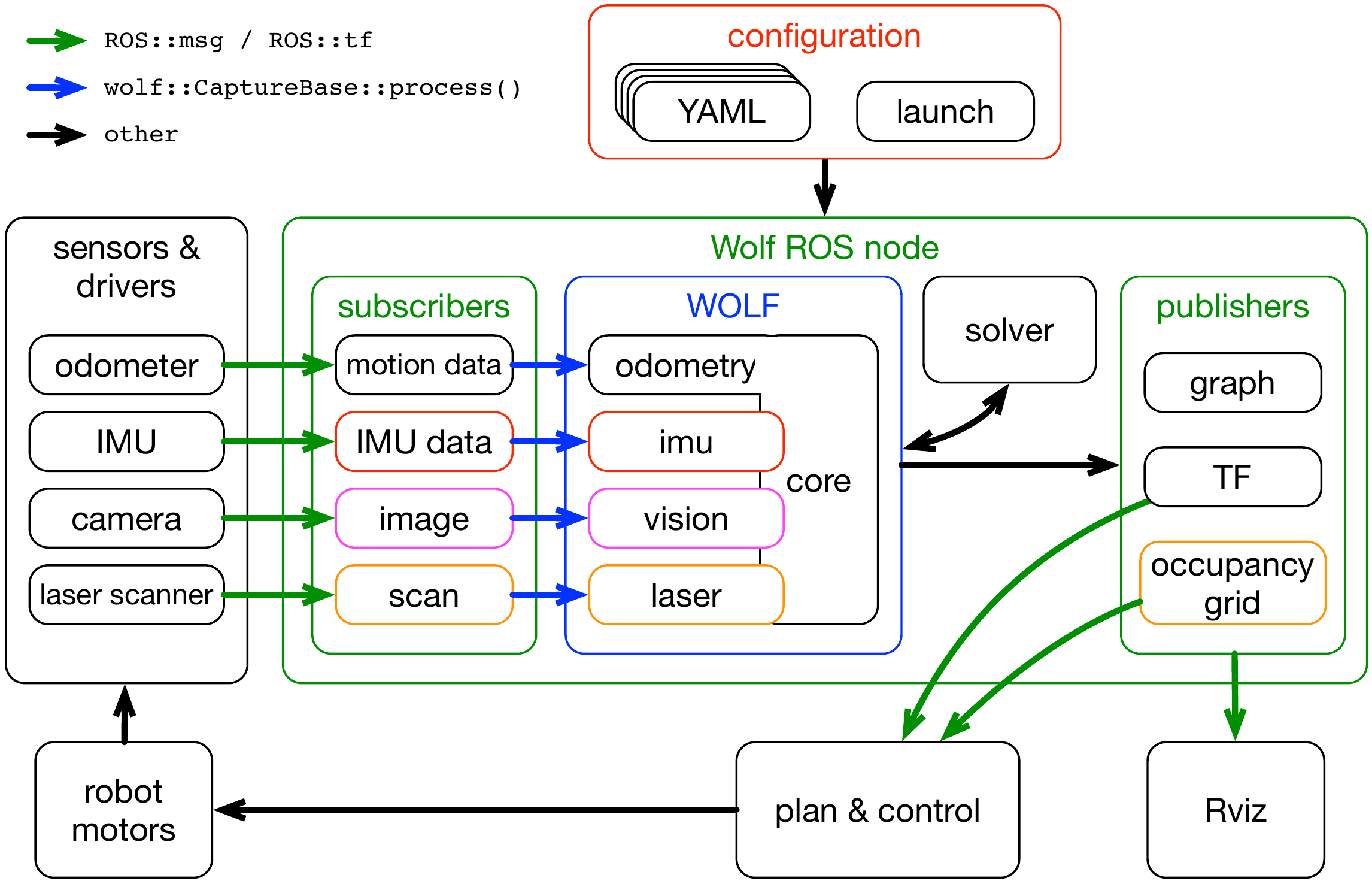}
  \caption{Using \ros\ to integrate a robotics estimation problem with \wolf. 
  Each sensor modality uses a \wolf\ plugin (colored boxes; odometry is part of \texttt{core}) and its corresponding \ros\ subscriber (in the same color). 
  Publishers create \ros\ messages with useful output data for control and visualization.}
  \label{fig:ROS-architecture}
\end{figure}

The \wolf\ integration in \ros\ is performed using a node and a set of subscribers and publishers.
We provide a unique generic \ros\ node and several \ros\ packages with subscribers and publishers following the same structure as the plugins.
A basic representation of the information flow in \wolf\ using \ros\ is sketched in \figRef{fig:ROS-architecture}.
{\rev
We also provide a built-in profiler to help users to optimize the performances in terms of accuracy and real-time constraints.
}

\subsection{The \wolf\ \ros\ node}

In the core \ros\ package, we provide a node that should serve many different robotics projects without modification.
The node has a list of publishers, a list of subscribers, the solver wrapper, and the \wolf\ tree.
Given a set of available plugins, all the user has to do is to write the YAML file(s) (see the accompanying video). 
These files specify the \wolf\ tree (see \ref{sec:autoconf}), and the subscribers and publishers. 
A straightforward \ros\ launch file specifying the YAML path completes the application.
At startup the node will: 
(a) parse the YAML configuration files;
(b) load the required \wolf\ plugins and initialize the \wolf\ tree;
(c) configure the solver;
and
(d) load and initialize all subscribers and publishers.

\subsection{\wolf\ subscribers and publishers}

A \wolf\ subscriber is an object that has a \ros\ subscriber and access to the corresponding \sensor\ in the \wolf\ tree.
Derived subscribers have to implement the appropriate callback method. 
This callback only has to create the derived \capture\ object and call its \texttt{process()} method.
\wolf\ takes care of the rest, launching all the \sensor's child \processors\ to process the \capture's data.

A \wolf\ publisher has a \ros\ publisher, access to the \wolf\ problem, and a user-defined publishing rate.
Derived publishers have to implement the method that fills a \ros\ message with the desired information from the \wolf\ tree, and publish it.
We provide in \texttt{core} publishers for the \ros\ transforms (tf) and the visualization of the graph. 


\section{Real cases}
\label{sec:experiments}

\begin{figure*}[t]
\centering
\begin{minipage}[t]{0.185\textwidth}
\includegraphics[width=\linewidth]{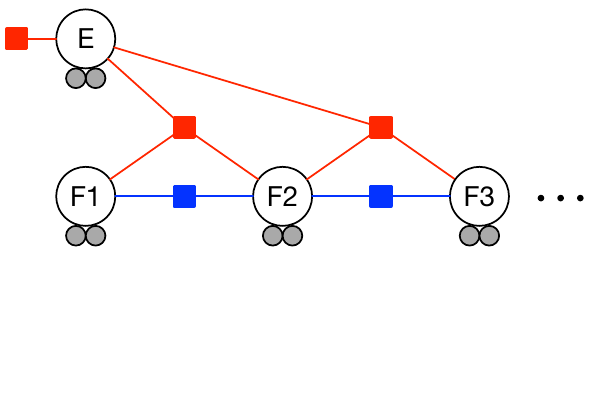}
\footnotesize{{(a) Demo
:} 2D LIDAR (red), wheel odometry (blue). 
 Fi:~frames position $p$ and orientation $q$; E:~LIDAR extrinsics $p,q$.}
\end{minipage}
\hfill
\begin{minipage}[t]{0.185\textwidth}
\includegraphics[width=\linewidth]{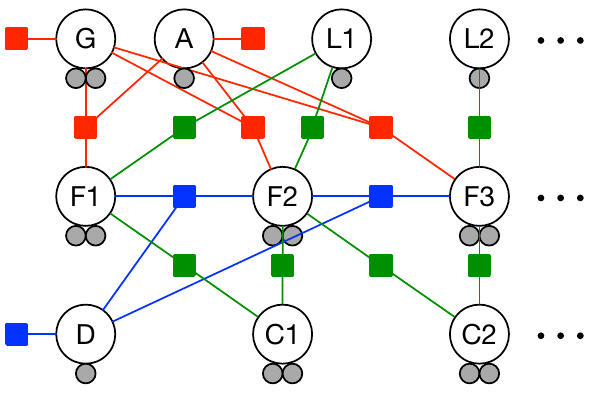}
\footnotesize{{(b) LOGIMATIC
:} GNSS (red), 2D LIDAR (green), differential drive (blue). G:~geo-referencing $p,q$; Li:~landmarks $p$; D: diff. drive; Ci:~Containers $p,q$.}
\end{minipage}
\hfill
\begin{minipage}[t]{0.185\textwidth}
\includegraphics[width=\linewidth]{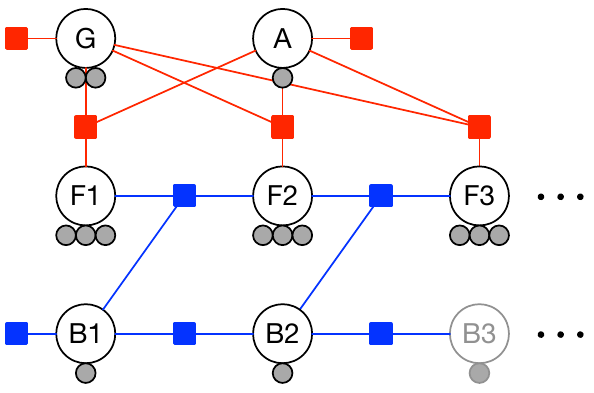}
\footnotesize{{(c) GAUSS
:} GNSS (red), IMU (blue). G:~geo-referencing $p,q$; A:~antenna $p$; Fi:~frames $p,q$ and velocity $v$; Bi:~IMU bias.}
\end{minipage}
\hfill
\begin{minipage}[t]{0.185\textwidth}
\includegraphics[width=\linewidth]{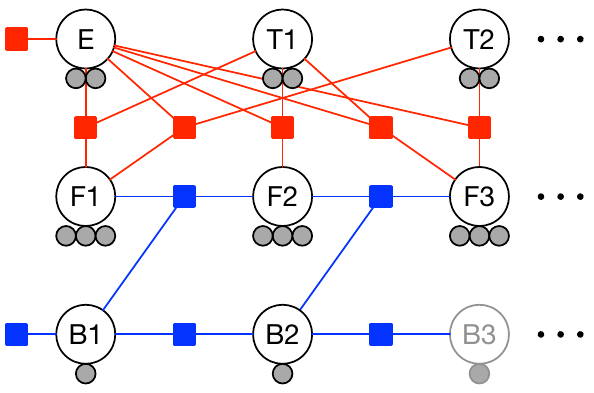}
\footnotesize{{(d) MEMMO-vision
:} Visual tags (red), IMU (blue). Ti:~visual tags $p,q$; E:~camera extrinsics $p,q$.}
\end{minipage}
\hfill
\begin{minipage}[t]{0.185\textwidth}
\includegraphics[width=\linewidth]{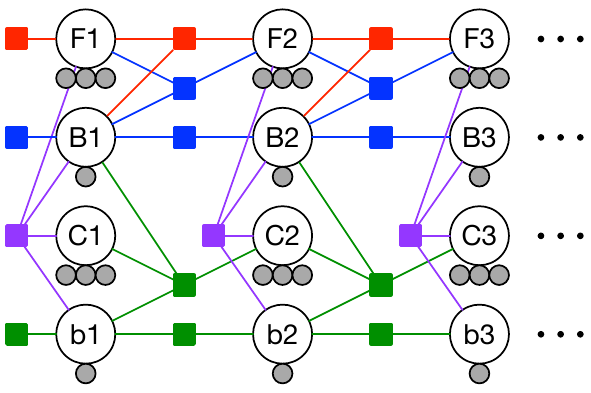}
\footnotesize{{(e) MEMMO-forces
:} IMU (blue), contact forces (green), joint angles (red and purple). Ci:~centroidal CoM $p,v$, angular momentum $L$; bi:~CoM bias.}
\end{minipage}
\caption{Human-readable graphs (see \figRef{fig:wolf-graph}) generated by \wolf\ for the five real cases exposed. Factors in colors (see sub-captions). \wolf\ nodes in labeled circles. State blocks in gray dots (see sub-captions). See \secRef{sec:experiments} and subsections therein for further details.  
}
\label{fig:graphs-real}
\end{figure*}

We present five distinct, varied, and illustrative applications of \wolf. 
They share significant parts of the code, re-use plugins, and are only determined by the set of YAML files parsed at launch time. 
All experiments run in real-time on standard modern PC hardware.
Please refer to \figRef{fig:graphs-real} for the human-readable factor graphs associated with each one of them.


\subsection{{\wolf\ demo:} 2D LIDAR + odometry (plugins: core and laser)}
\label{sec:demo}

\begin{figure*}[tb]
\centering
\begin{minipage}[t]{\columnwidth}
\centering
\includegraphics[height=0.54\columnwidth, trim=30 0 0 0, clip]{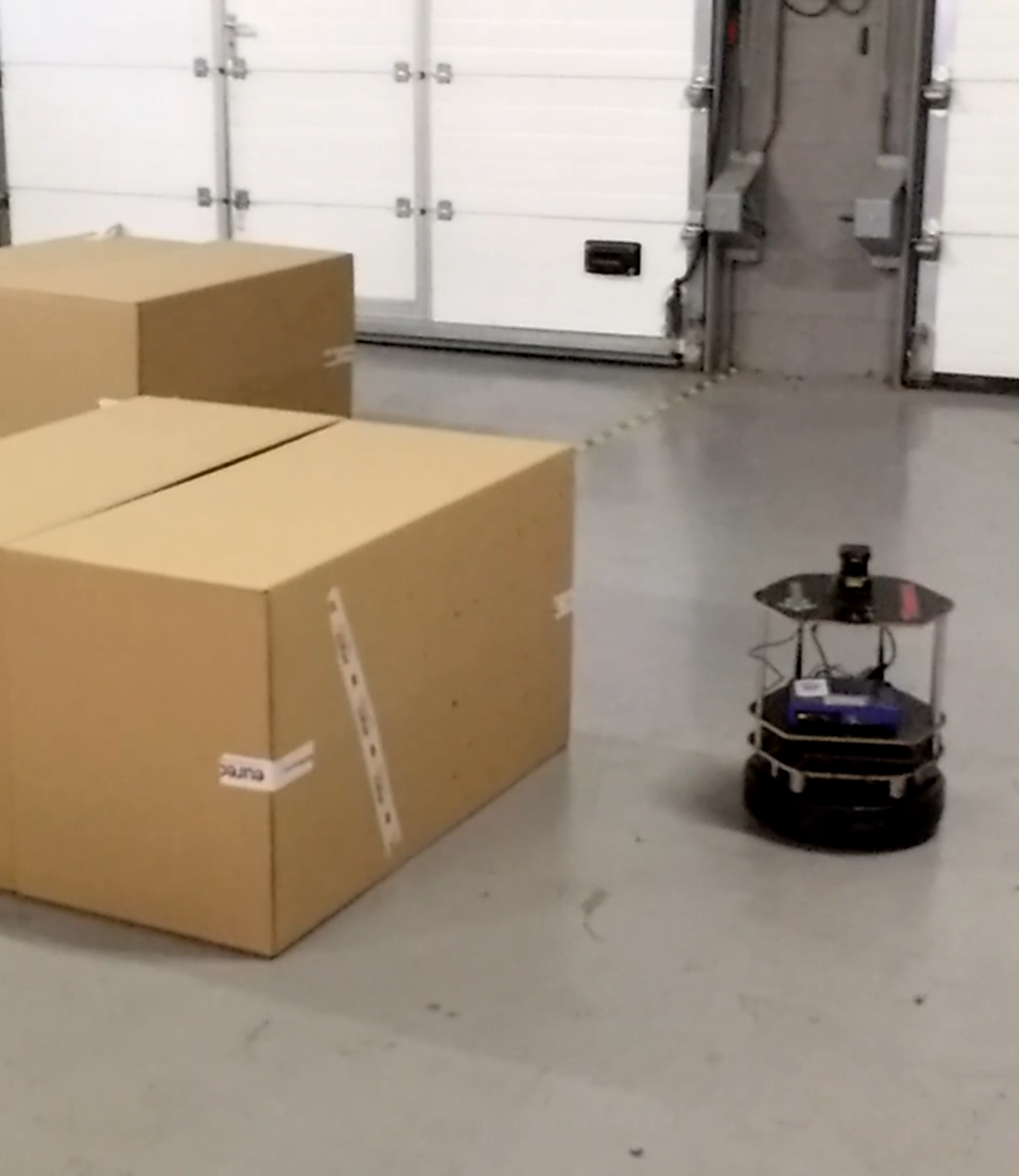}
\hfill
\includegraphics[height=0.54\columnwidth, trim=150 0 702 0, clip]{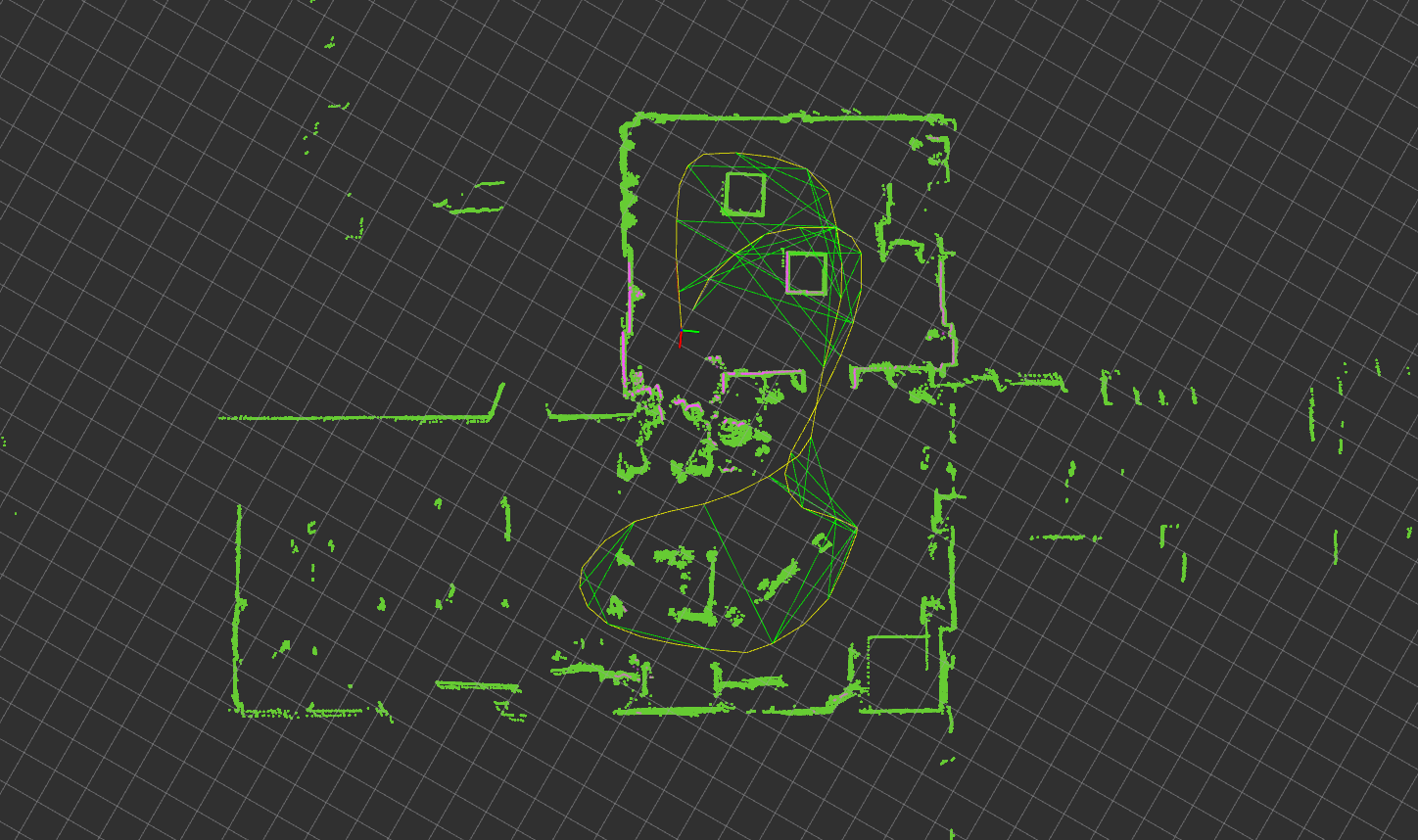}\!\!\!
\includegraphics[height=0.54\columnwidth, trim=377 0 510 0, clip]{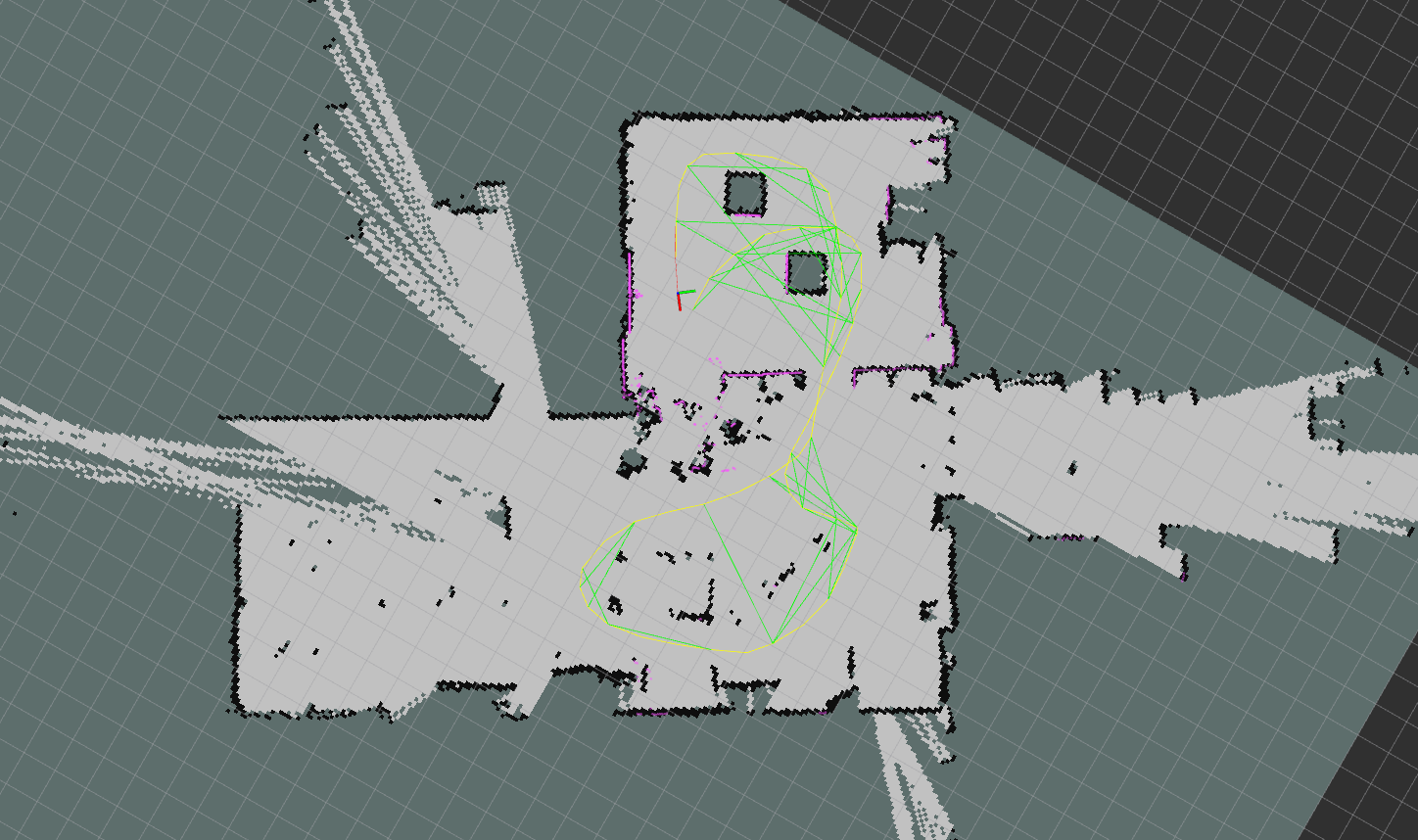}\!\!\!
\includegraphics[height=0.54\columnwidth, trim=569 0 320 0, clip]{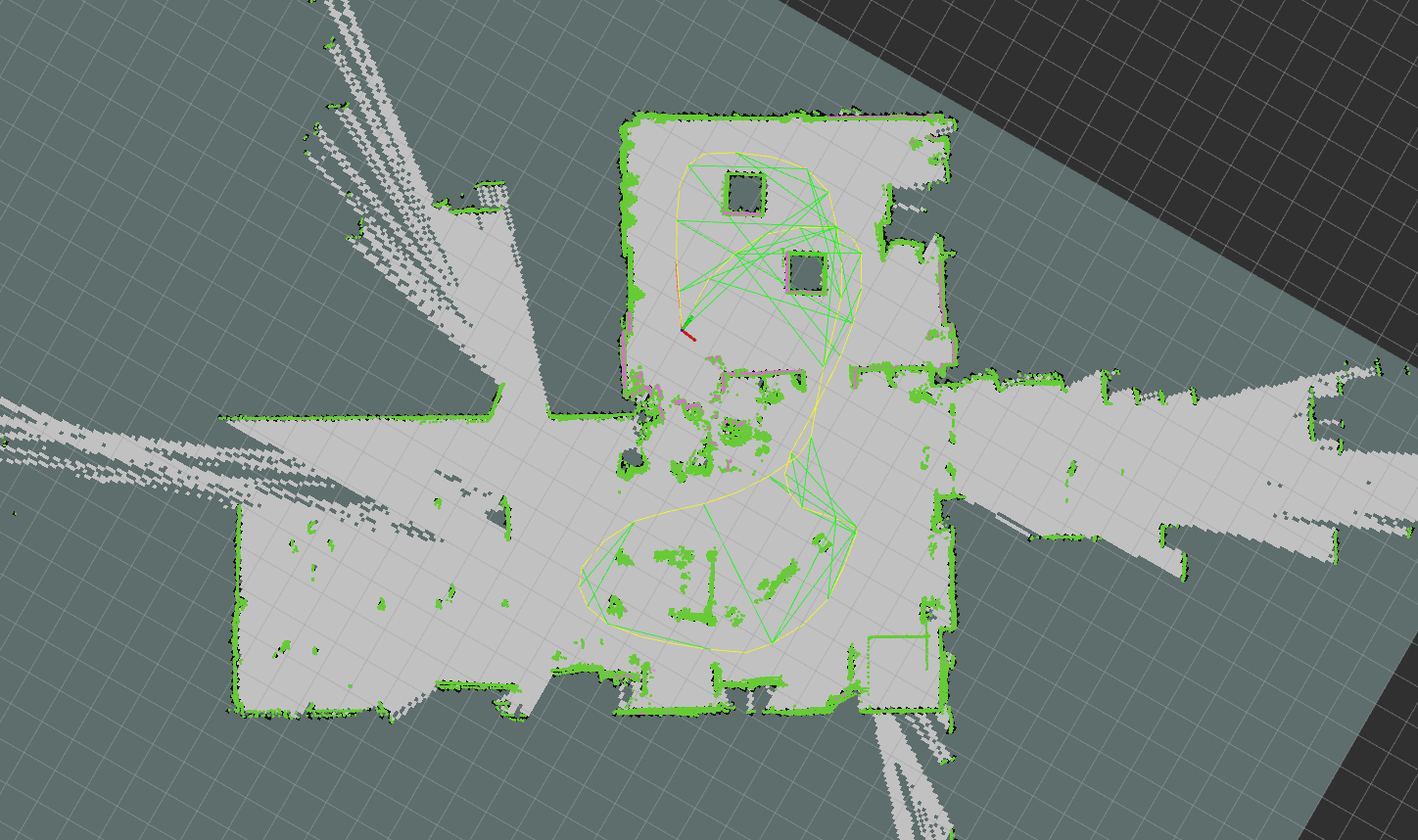}
\caption{A turtlebot equipped with a low cost LIDAR and wheel odometry performs indoor localization and mapping. This toy application constitutes a tutorial for getting started with \wolf\ and \ros, and is shipped with the library. 
The map can be published as pointcloud (left fraction of the map), as occupancy grid (center), or both (right).}
\label{fig:demo}
\end{minipage}
\hfill
\begin{minipage}[t]{\columnwidth}
\centering
\includegraphics[height=0.54\linewidth]{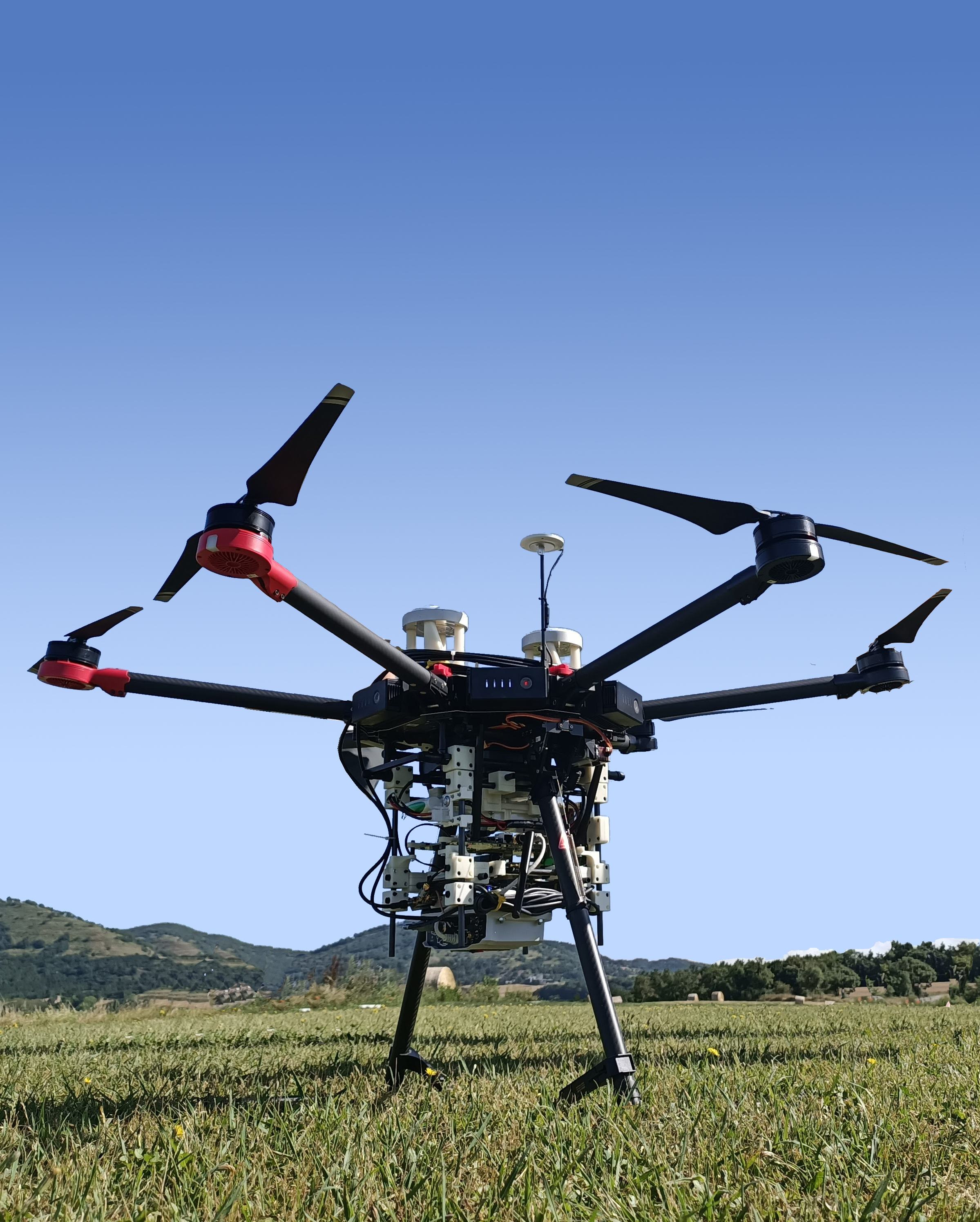}
\quad
\includegraphics[height=0.54\linewidth]{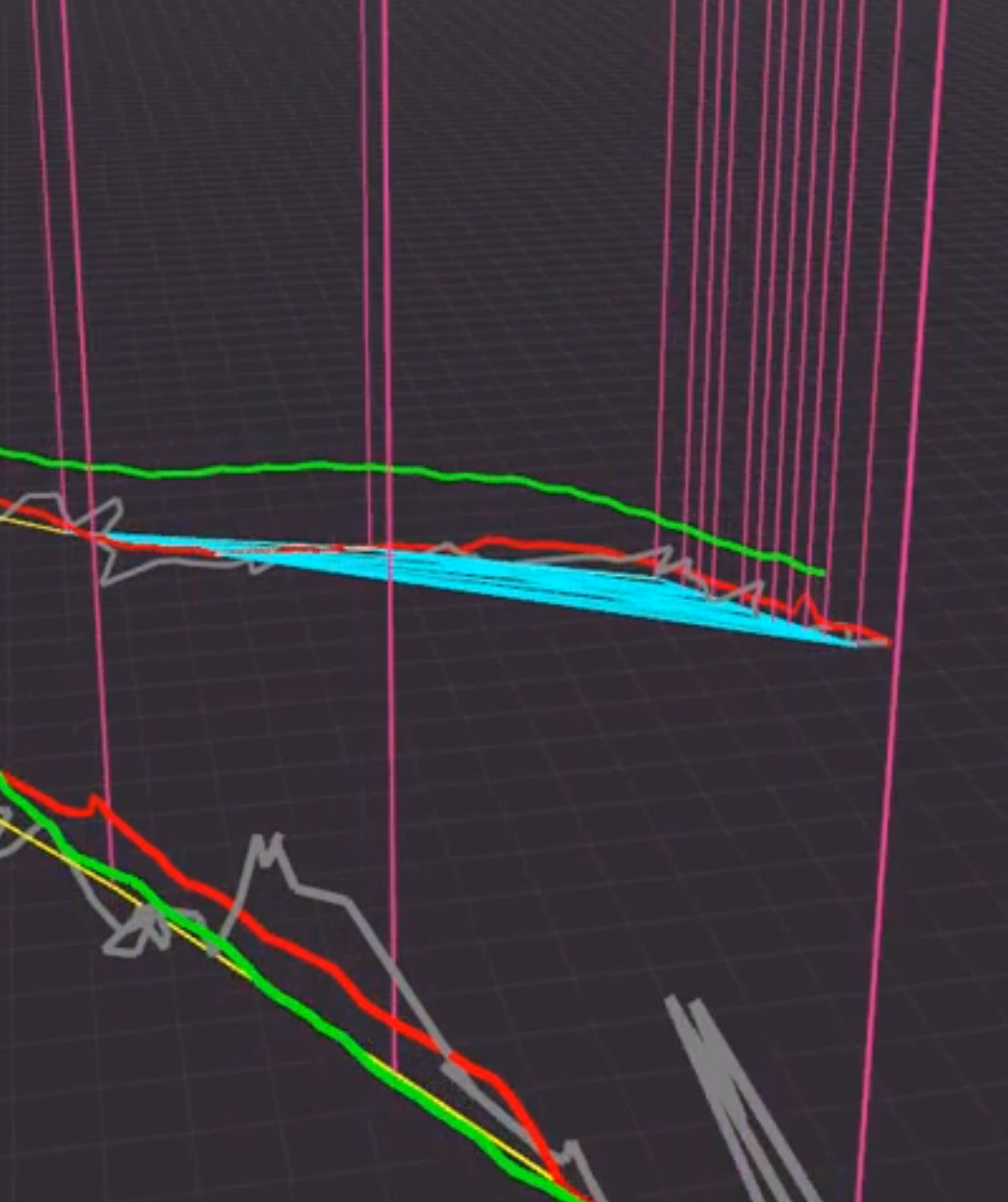}
\caption{In the context of the EU H2020 project GAUSS, we equipped a UAV with two GNSS receivers and an IMU. Plain GNSS fixes show numerous multi-path artifacts yielding an erratic trajectory (gray). We combined IMU, GNSS pseudoranges (magenta) and TDCP (cyan) together with robust outlier rejection to achieve smooth and accurate motion estimation (red).}
\label{fig:gnss}
\end{minipage}
\end{figure*}

This application involves a differential-drive base with a 2D laser scanner (\figRef{fig:graphs-real} (a), \figRef{fig:demo}). 
It exhibits: a 2D odometry pre-integrator; a laser processor performing laser odometry and mapping based on scan matching, with self-calibration of LIDAR extrinsics; and a simple loop closer also based on scan matching. 
Frames are produced by the laser odometry processor according to a mixed policy based on distance traveled, angle turned, and the quality of the scan matching. 
The system, targeted at educational tasks in robotics, can navigate small indoor areas, publishing the graph and the map in two possible formats: a point-cloud that may be used for visualization and debugging, and an occupancy grid useful for planning and control. 
It constitutes one of the \wolf-\ros\ demos shipped with the library.

\subsection{{LOGIMATIC:} GNSS + differential drive odometry + LIDAR (plugins: core, gnss and laser)}
\label{sec:logimatic}

\begin{figure*}[tb]
\centering
\includegraphics[height=0.28\linewidth]{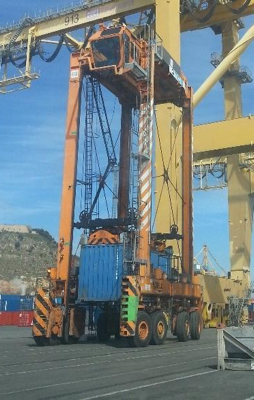}
\hfill
\includegraphics[height=0.28\linewidth]{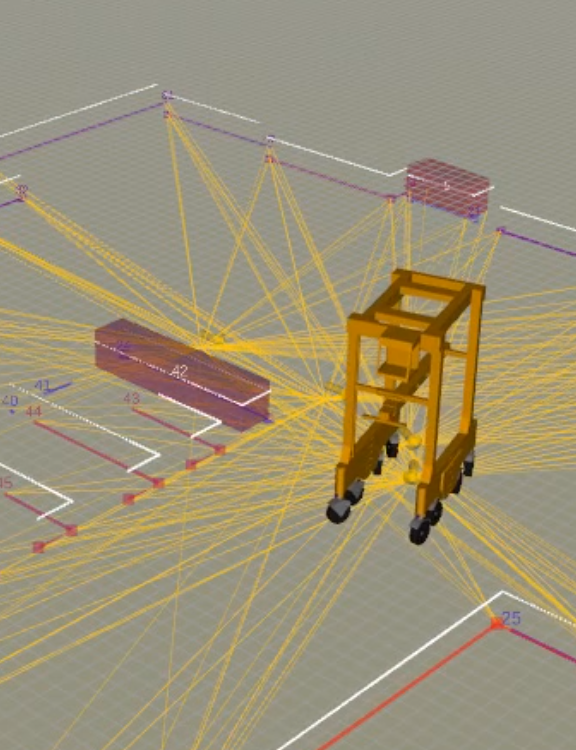}
\hfill
\includegraphics[angle=90,height=0.28\linewidth]{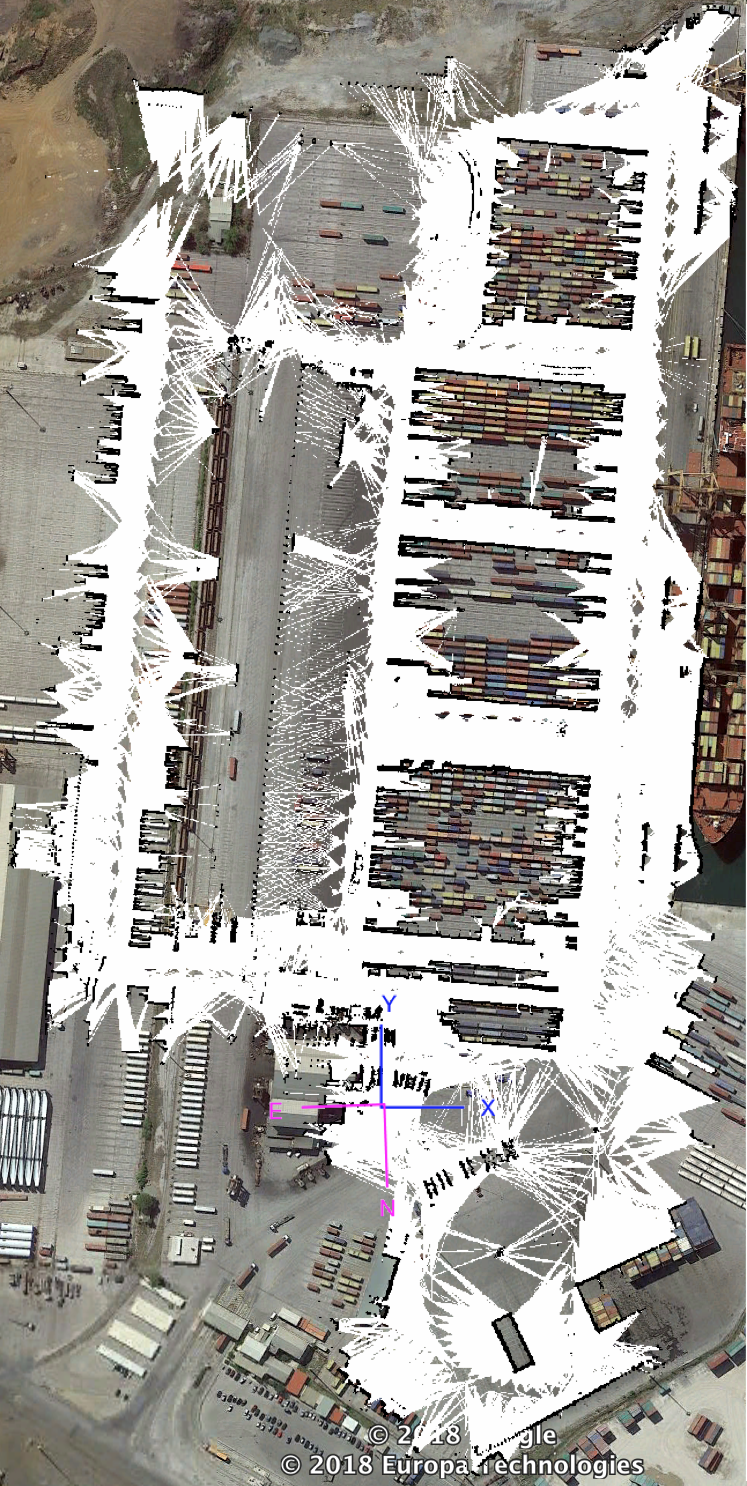}
\caption{In the context of the EU H2020 LOGIMATIC project, an autonomous retrofitted 8-wheel straddle carrier equipped with GNSS, LIDAR and odometry (left) mapped an area and detected the containers (center) in the port of Thessaloniki in Greece. The final map produced with \wolf\ has an area of about 1000m$\times$500m and is shown overlaid on top of an aerial photograph (right). The map reference frame (blue, XY) was precisely geo-referenced by \wolf\ \wrt the local ENU coordinates (magenta, EN) {\rev established} by the first GNSS fix.}
\label{fig:2D-maps}
  \vspace{-8pt}
\end{figure*}

This application performs 2D SLAM in a large port area (\figRef{fig:graphs-real} (b), \figRef{fig:2D-maps}) \cite{rizzo-19-logimatic}. 
A big 8-wheeled straddle-carrier is equipped with one GNSS, left and right wheel groups rotary encoders, and 4 long-range LIDARs.
A motion processor integrates wheel odometry. 
One GNSS processor incorporates GNSS fixes as absolute factors at frames. 
A second GNSS processor computes globally referenced displacements from satellite time-differenced carrier phase (TDCP) accurate to the cm level~\cite{graas-04-tdcp,freda-15-tdcp}. 
Four cooperative LIDAR processors extract and track poly-line features, used for SLAM, and can detect containers and place them in the SLAM map as objects.
The system also performs precise geo-localization of the vehicle and mapped area, and self-calibrates the differential drive motion model \cite{DERAY-19-SELFCALIB} and the antenna location.

\subsection{{GAUSS:} GNSS + IMU (plugins: core, imu and gnss)}
\label{sec:gnss}

This 3D implementation involves a UAV with a GNSS receiver, and a low-cost IMU (\figRef{fig:graphs-real} (c), \figRef{fig:gnss}) \cite{Jmenez2019GalileoAE}. 
We use an IMU pre-integrator with bias tracking. 
GNSS is incorporated in a tightly coupled manner via individual pseudo-ranges as frame-to-satellite factors.
The same GNSS processor also adds individual TDCP factors between pairs of frames at up to 1\,min interval.
Given a sufficient number of satellites, direct pseudoranges allow for absolute positioning in the 1m range, while TDCP observes absolute displacements in the cm range, much more accurate than individual fixes.
The antenna location is self-calibrated.

\begin{figure*}[tb]
\centering
\vspace{1.5mm}
\begin{minipage}[t]{1.15\columnwidth}
\centering
\includegraphics[height=0.28\linewidth]{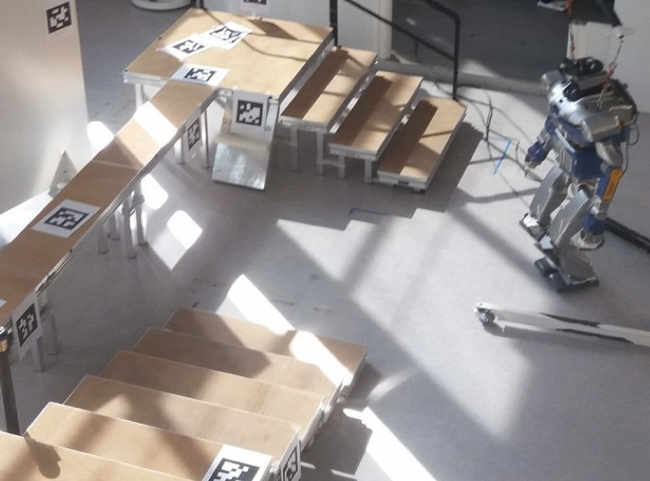}
\hfill
\includegraphics[height=0.28\linewidth, trim={1.2cm 0 0 0}, clip]{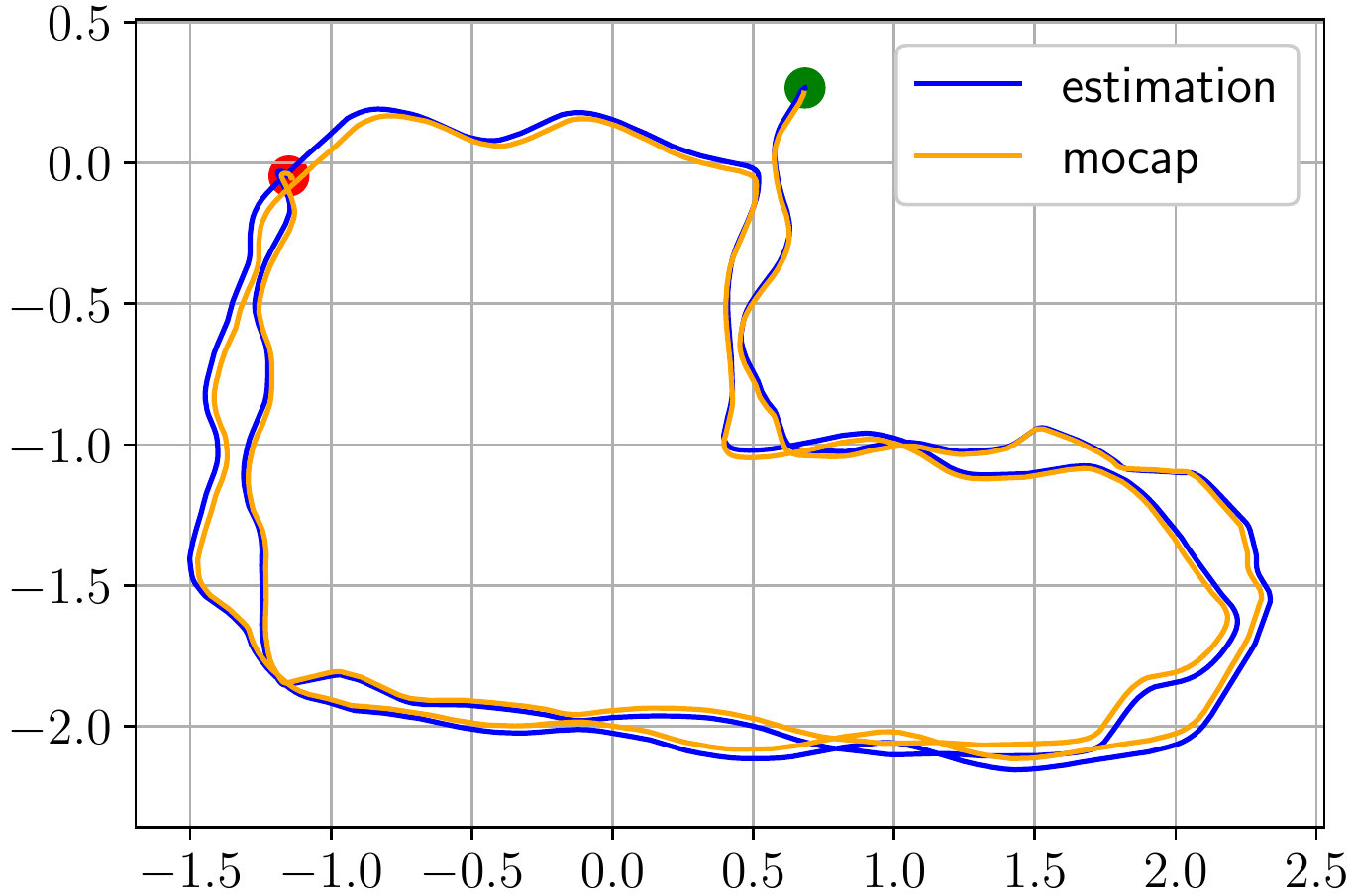}
\hfill
\includegraphics[height=0.28\linewidth, trim={5cm 0 6cm 0}, clip]{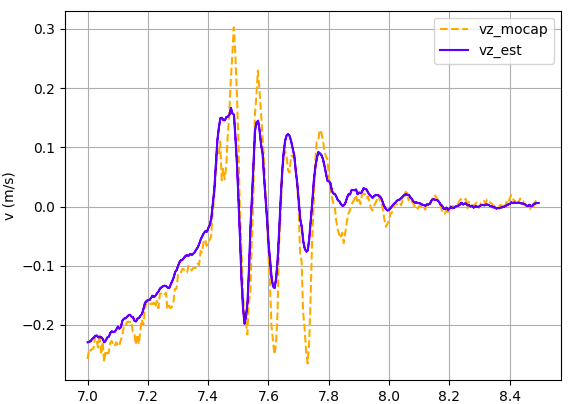}
\caption{In the context of the EU H2020 MEMMO project, the humanoid HRP-2 with a visual-inertial setup on its head performs a loop on a circuit with visual tags (left) at LAAS-CNRS. We show a drift-free trajectory reconstruction (center; "mocap" = IR motion capture) and the recovery, at the IMU sampling rate of 200Hz and in the same global reference, of head vibrations (right) resulting from foot impacts while descending the stairs.}
\label{fig:humanoid}
\end{minipage}
\hfill
\begin{minipage}[t]{0.85\columnwidth}
\centering
\includegraphics[height=0.37\linewidth]{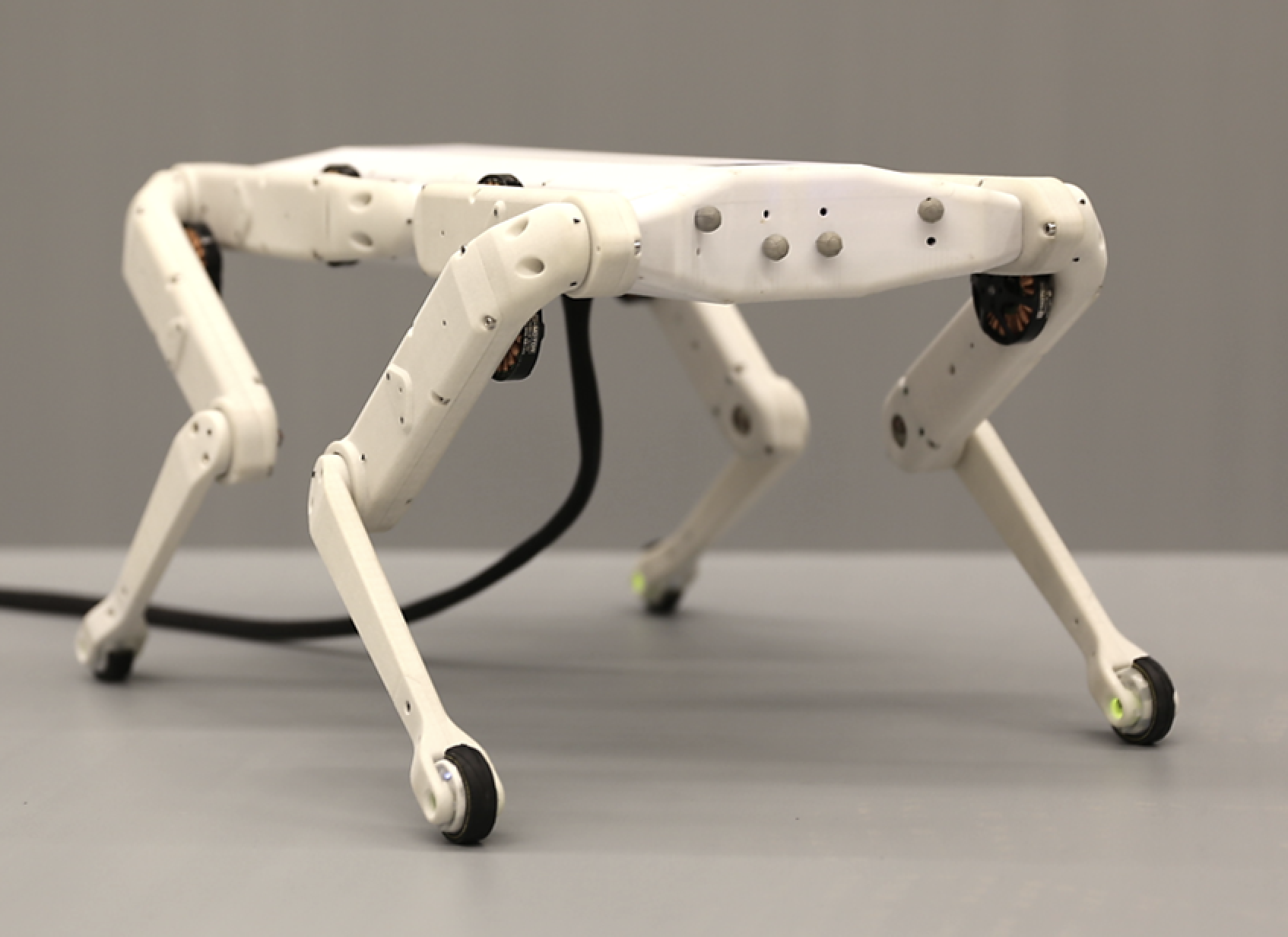}
\hfill
\includegraphics[height=0.37\linewidth]{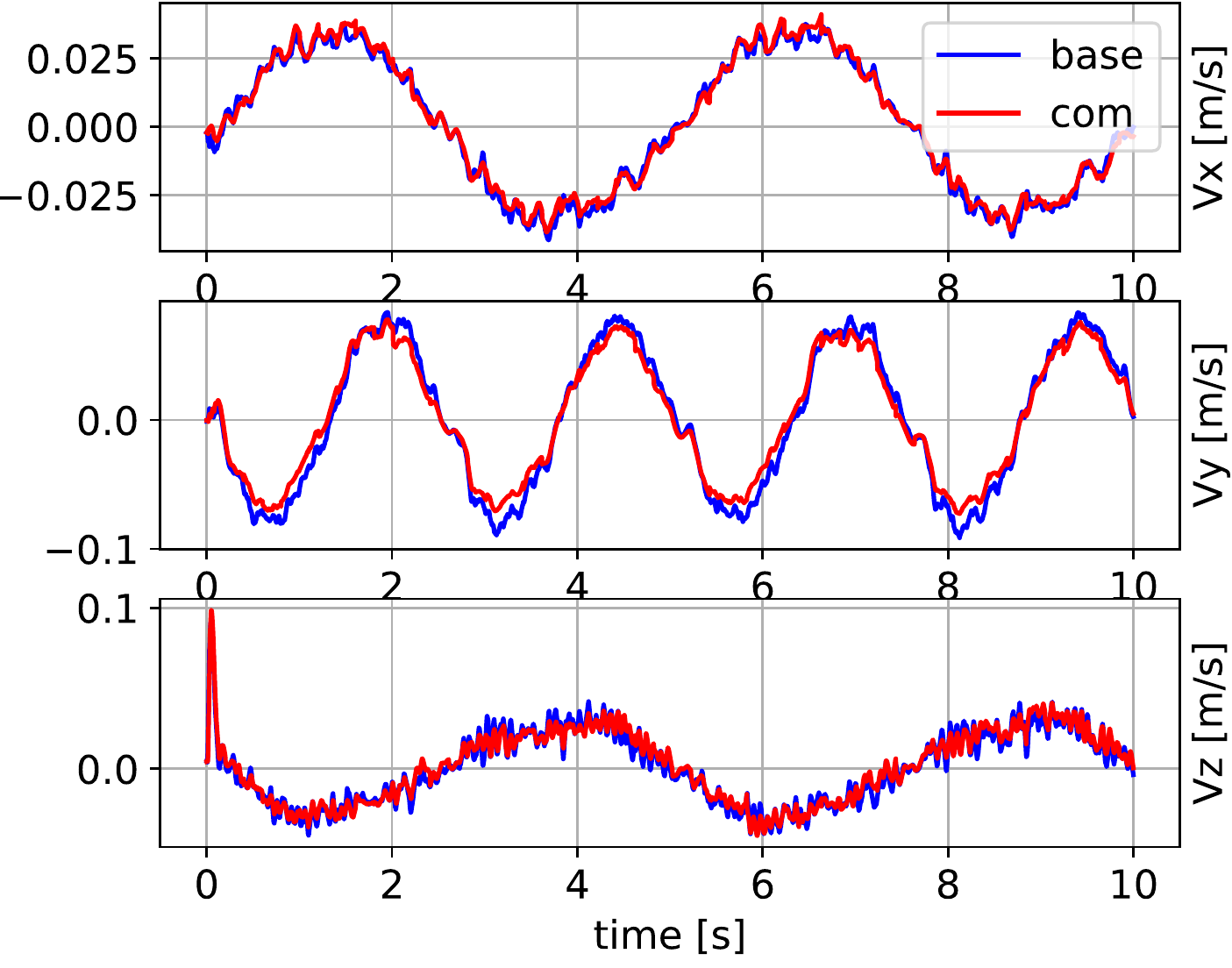}
\caption{In the context of the EU H2020 MEMMO project, the quadruped Solo12 (left) with measurements of an IMU, limb encoders and motor currents is used for whole-body base and centroidal estimation. We show precise recovery at 1kHz of the base position, the center of mass and its velocity.}
\label{fig:solo}
\end{minipage}
\end{figure*}

\subsection{{MEMMO-vision:} 6DoF vision + IMU for the humanoid (plugins: core, imu, vision and apriltag)}
\label{sec:humanoid}

This 3D experiment (\figRef{fig:graphs-real} (d), \figRef{fig:humanoid}) involves a humanoid robot equipped with a camera and an IMU in its head. 
The IMU processor is the same as in \ref{sec:gnss}. 
The visual processor detects AprilTags \cite{olson-11-apriltag} in the scene, measuring 6~DoF camera-tag transforms. 
The tag locations can be either known a-priori (map-based localization), completely unknown (SLAM), or a mixture of them (SLAM with prior visual clues). 
All these variants are selected at YAML level.

\wolf\ produces drift-free trajectory estimates accurate to $\le4$\,cm.
It also publishes full robot states referred to a precise gravity direction at the IMU frequency of 200Hz~\cite{fourmy-19-april}. 

In a similar approach, we conceived an object-based visual-inertial SLAM~\cite{debeunne-21-icra-hal} (plugins: core, imu, vision and objectslam), where we use the deep learning object detection library CosyPose~\cite{labbe-20-cosypose-eccv} instead of AprilTags.
CosyPose identifies previously learned objects and returns the 6DoF camera-object relative pose.
The factor graph is thus the same as in the AprilTag system (\figRef{fig:graphs-real} (d)).

\subsection{{MEMMO-forces:} IMU, contact forces and joint angles for the quadruped (plugins: core, imu and bodydynamics)}
\label{sec:solo}

The last application of \wolf\ showcased in this paper performs whole-body estimation~\cite{fourmy-21-centroidal} of the quadruped Solo12 from the Open Dynamic Robot Initiative (\figRef{fig:graphs-real} (e), \figRef{fig:solo}).
Contact forces altering the centroidal robot dynamics (position and velocity of the center of mass (CoM), angular momentum) are pre-integrated using a specialization~\cite{fourmy-21-centroidal} of our generalized pre-integration method (\secRef{sec:proc_motion}). 
The IMU data is also pre-integrated to measure the motion of the base. 
Joint angle measurements are then participating in the estimation in two ways: by providing leg odometry displacements, contributing to the observability of IMU biases; and by relating the base and centroidal states, thus connecting the whole graph and producing a tightly-coupled whole-body estimator.
The states of base position, orientation and velocity, IMU biases, CoM position and velocity, angular momentum and CoM sensor bias are published at 1\,kHz with a lag of less than 1\,ms.
This provides accurate observation of the centroidal dynamics, which is crucial for balance and gait control.


\section{Conclusions and future work}
\label{sec:conclusions}

Over the last six years, \wolf\ has provided a consistent framework for conceiving and implementing sensor fusion for a vast variety of research projects.
We have used it to incorporate abstract algorithms and this has pushed us to truly understand the common connections between many different robotics estimation problems.
\wolf\ has matured during this time and we believe it offers a competitive design, with many interesting features that are desirable in such kind of software packages.
We now offer it open to the community with the hope that it can be useful to other teams.

Work needs to be done to improve the real-time performance. 
In particular, \wolf\ still lacks the possibility to launch each processor in an individual thread.
This is a weak point that is common to many other estimation frameworks in the state of the art, and it seems not straightforward to solve.
We believe that effort must be made in this direction, either in \wolf\ and/or elsewhere to be able to bring modular, tightly coupled and reconfigurable multi-sensor fusion to reality.


%
%

\section*{Acknowledgments}

The authors would like to thank 
Dylan Bourgeois, 
C\'esar Debeunne, 
Idril Geer, 
Peter Guetschel, 
Josep Mart\'i-Saumell, 
Sergi Pujol, 
\`Angel Santamaria-Navarro, 
Jaime Tarraso, 
Pier Tirindelli 
and 
Oriol Vendrell
for their valuable contributions to this project.

\bibliographystyle{./bibliography/IEEEtran}
\bibliography{./bibliography/bibJuan.bib} 

\end{document}